\documentclass[conference]{IEEEtran}
\IEEEoverridecommandlockouts

\usepackage{hyperref}
\usepackage{cite}
\usepackage{amsmath,amssymb,amsfonts, amsthm}
\usepackage{bbding}
\usepackage{algorithmic}
\usepackage{graphicx}
\usepackage{multicol}
\usepackage{multirow}
\usepackage{textcomp}
\usepackage{units}
\usepackage{booktabs}
\usepackage{array}
\usepackage{makecell}
\usepackage{xcolor}
\newcolumntype{M}[1]{>{\centering\arraybackslash}m{#1}}
\newtheorem{theorem}{Theorem}
\newtheorem{definition}{Definition}

\def\BibTeX{{\rm B\kern-.05em{\sc i\kern-.025em b}\kern-.08em
    T\kern-.1667em\lower.7ex\hbox{E}\kern-.125emX}}
\begin{document}

\title{Detecting OOD Samples via Optimal Transport Scoring Function
}

\author{
\IEEEauthorblockN{
Heng Gao}
\IEEEauthorblockA{\textit{ISTBI} \\
\textit{Fudan University}\\
Shanghai, China \\
hgao22@m.fudan.edu.cn
}
\and
\IEEEauthorblockN{
Zhuolin He
}
\IEEEauthorblockA{\textit{School of Computer Science} \\
\textit{Fudan University}\\
Shanghai, China \\
zlhe22@m.fudan.edu.cn}
\and
\IEEEauthorblockN{
Jian Pu$^*$\thanks{*Corresponding author.}}
\IEEEauthorblockA{\textit{ISTBI} \\
\textit{Fudan University}\\
Shanghai, China \\
jianpu@fudan.edu.cn}
}

\maketitle

\begin{abstract}
To deploy machine learning models in the real world, researchers have proposed many OOD detection algorithms to help models identify unknown samples during the inference phase and prevent them from making untrustworthy predictions. Unlike methods that rely on extra data for outlier exposure training, post hoc methods detect Out-of-Distribution (OOD) samples by developing scoring functions, which are model agnostic and do not require additional training. However, previous post hoc methods may fail to capture the geometric cues embedded in network representations. Thus, in this study, we propose a novel score function based on the optimal transport theory, named OTOD, for OOD detection. We utilize information from features, logits, and the softmax probability space to calculate the OOD score for each test sample. Our experiments show that combining this information can boost the performance of OTOD with a certain margin. Experiments on the CIFAR-10 and CIFAR-100 benchmarks demonstrate the superior performance of our method. Notably, OTOD outperforms the state-of-the-art method GEN by $7.19\%$ in the mean FPR@95 on the CIFAR-10 benchmark using ResNet-18 as the backbone, and by $12.51\%$ in the mean FPR@95 using WideResNet-28 as the backbone. In addition, we provide theoretical guarantees for OTOD. The code is available in \url{https://github.com/HengGao12/OTOD}.
\end{abstract}


\section{Introduction}
To prevent machine learning models from producing untrustworthy predictions and being overconfident on Out-of-Distribution (OOD) inputs \cite{b1}, various OOD detection algorithms have been proposed to enable models to distinguish whether a test sample comes from in-distribution (ID) data or unseen domains. 

The post hoc-based methods identify OOD samples by developing scoring functions, which are free of training and model agnostic. In MSP \cite{b2}, Hendrycks et al. first propose a strong baseline that utilizes the statistics of softmax probability outputs to identify outliers. ODIN \cite{b10} proposes detecting OOD samples by adding input perturbation and temperature scaling. In EBO \cite{b3}, the authors develop an energy-based score function that uses information from the model's logits outputs, surpassing many softmax-based scores and generative-based methods. In recent work, GEN \cite{b4} proposes using generalized entropy to calculate the OOD scores of test samples. However, limited attention is paid to study how to develop a score function based on optimal transport distances, which possess good theoretical properties \cite{b7} and are capable of capturing the spatial discrepancies between probability distributions \cite{b23}.

\begin{figure}[t]
\centerline{\includegraphics[width=0.92\linewidth]{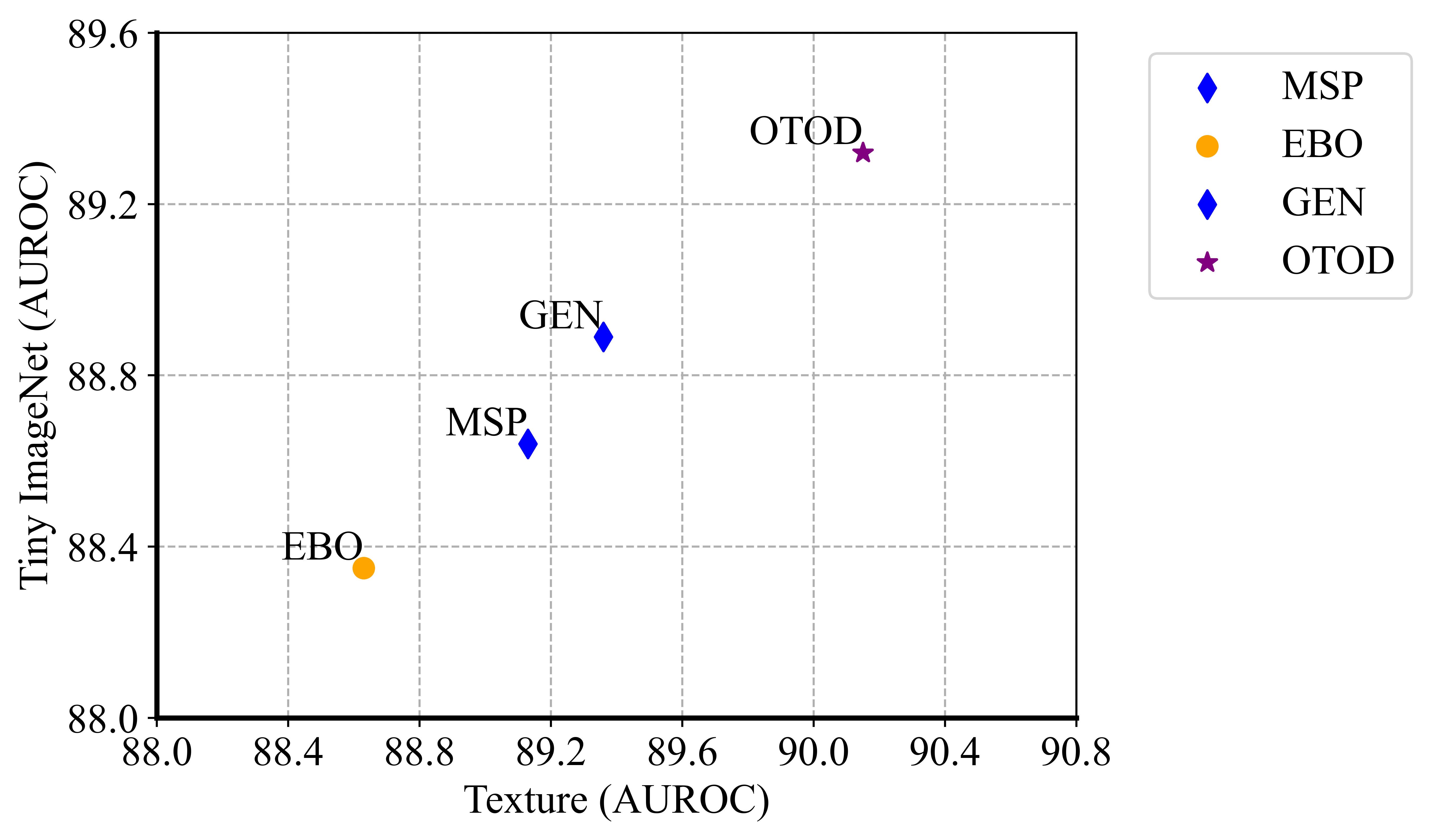}}
\caption{The AUROC (in percentage) of four OOD detection methods using ResNet-18 \cite{b5} trained on CIFAR-10 \cite{b6}. The OOD datasets are Tiny ImageNet (TIN) \cite{b11} and Texture \cite{b14}. Methods marked with blue $\lozenge$ use the softmax probability input; methods marked with orange $\circ$ use the logits inputs. Our proposed method OTOD (marked with purple $\star$) uses information from logits, softmax probability, and features.}
\label{fig-1}
\end{figure}

 Therefore, in this paper, we develop a post hoc method that leverages the optimal transport distance \cite{b7}, also known as Wasserstein distance, to calculate OOD scores without any training techniques or modifying the network architecture. Specifically, we first employ Wasserstein-1 distance to calculate OOD scores using features extracted from the penultimate layer of the pre-trained neural network. Unlike methods that utilize statistical distances, such as MDS \cite{b8} and KLM \cite{b9}, OTOD does not require estimating the empirical class mean and covariance for the ID data during the OOD testing phase. Then, we combine information from the logits and the softmax probability space to calculate OOD scores. We find that integrating this information helps to improve the OOD detection results. Moreover, we apply the temperature scaling technique \cite{b10} to further enhance the OOD detection performance. The experiments on the CIFAR-10 and CIFAR-100 \cite{b6} benchmarks demonstrate the superior performance of OTOD, which outperforms many state-of-the-art post hoc methods. Furthermore, we provide theoretical guarantees for OTOD.

The main contributions of our paper are: (1) We develop a novel post hoc method based on optimal transport theory, named OTOD, to calculate OOD scores of test samples without using any in-distribution data. (2) We find that the combination of features, logits, and softmax probability inputs allows OTOD to achieve non-trivial improvements in several commonly used metrics (the FPR@95 and AUROC) on the CIFAR-10/100 \cite{b6} benchmarks. 
(3) Both empirical and theoretical results demonstrate the superior performance of OTOD.

\section{Background}  


\subsection{Wasserstein Distances}
Now we introduce the definition of Wasserstein distances.

\begin{definition}[Wasserstein distances \cite{b7}]\label{w1-distance}
Denote $\left(\mathbf{X}, d\right)$ as a Polish metric space and let $p\in\left[1, \infty\right)$. For any two probability measures $\mu, \nu$ on $\mathbf{X}$, the Wasserstein distance of order $p$ between $\mu$ and $\nu$ is defined as follows:
\begin{equation*}
\mathcal{W}_{p}(\mu, \nu) =\left(\inf _{\pi \in \Pi(\mu, \nu)} \int_{\mathbf{X}} d(x, y)^{p} d \pi(x, y)\right)^{1 / p},
\end{equation*}
where $\Pi(\mu, \nu)$ is the set of all joint probability measures on $\mathbf{X}\times\mathbf{X}$ whose marginals are $\mu$ and $\nu$.
\end{definition}

Particularly, let $p=1$, we can obtain the $\mathcal{W}_1$ distance, also known as Kantorovich–Rubinstein distance, which can be written in the following form, 
\begin{equation*}
    \mathcal{W}_{1}(\mu, \nu)=\sup _{\|\psi\|_{\operatorname{L}} \leq 1}\left\{\int_{\mathbf{X}} \psi d \mu-\int_{\mathbf{X}} \psi d \nu\right\},
\end{equation*}
where $\psi\in L^1(\mu)$ is c-convex and 1-Lipschitz, $\vert\vert\cdot\vert\vert_{\operatorname{L}}$ denotes the Lipschitz norm of function $\psi$.



\section{Optimal Transport Scoring Function}

The main idea of our method is to develop a score function based on Optimal Transport theory \cite{b7} to calculate the OOD scores for test samples. Our method consists of two parts: (\romannumeral 1) Wasserstein-1 score calculation (in Section \ref{ot-cal}); (\romannumeral 2) multi-scale information integration (in Section \ref{ms-ii}). We also provide theoretical guarantees for OTOD partially (in Section \ref{theory}).

\subsection{Optimal Transport Score Calculation}\label{ot-cal}
For each input image $\mathbf{x}\in \mathbf{X}$, $\mathbf{X}$ is the input space, the feature of the given sample is denoted by $\mathbf{f}\in\mathbb{R}^{d}$. 
Then, we normalize feature $\mathbf{f}$ by using $L^2$ normalization. Denote the normalized feature as 
$
    \hat{\mathbf{f}} = \nicefrac{\mathbf{f}}{\vert\vert \mathbf{f}\vert\vert_2}
$.
In this case, $\hat{\mathbf{f}}$ can be approximated as a d-dimensional distribution.

Afterwards, we use $\mathcal{W}_1$ distance to calculate the OOD score for each test sample. To be specific, the Wasserstein-1 OOD score for sample $\mathbf{x}$ takes the form as
\begin{equation}
    \hat{S}_{\mathcal{W}_1}(\hat{\mathbf{f}}) = -\min\left[\mathcal{W}_1 (\hat{\mathbf{f}}, \mathbf{u}),  \mathcal{W}_1 (\hat{\mathbf{f}}, \mathbf{o})\right], \label{eq-1}
\end{equation}
where 
$\mathbf{u}$ is a mean vector, $\mathbf{o}$ is a zero vector. Note that, here we have a model ensemble effect by using both the mean vector and the zero vector to calculate the Wasserstein-1 score.

\subsection{Multi-scale Information Integration}\label{ms-ii}
To enhance the performance of OTOD, we further integrate the information contained in logits and softmax probability. Specifically, denote the logits of the given image sample as $\mathbf{l}\in\mathbb{R}^K$, $K$ is the number of classes. Denote the softmax probability as $\mathbf{p}\in[0, 1]^K$. Then, we also normalize vector $\mathbf{l}, \mathbf{p}$ and perform the same calculation as equation (\ref{eq-1}). Likewise, we denote the normalized logits and softmax probability as $\hat{\mathbf{l}}, \hat{\mathbf{p}}$. Therefore, the whole Wasserstein-1 score can be written as
\begin{equation}
    \Tilde{S}_{\mathcal{W}_1}(\hat{\mathbf{f}}, \hat{\mathbf{l}}, \hat{\mathbf{p}}) = \alpha_1\hat{S}_{\mathcal{W}_1}(\hat{\mathbf{f}}) + \alpha_2\hat{S}_{\mathcal{W}_1}(\hat{\mathbf{l}}) + \alpha_3\hat{S}_{\mathcal{W}_1}(\hat{\mathbf{p}}), \label{eq-3}
\end{equation}
where
\begin{equation}
    \sum\limits_{i=1}^3 \alpha_i = 1, \alpha_i\in [0,1]. \label{eq-4}
\end{equation}


Moreover, we adapt the temperature scaling technique from ODIN \cite{b10} to OTOD to further enlarge the discrepancy between ID and OOD softmax probability. Hence, the final score takes the form of the following:
\begin{equation}
    S_{\mathcal{W}_1}(\hat{\mathbf{f}}, \hat{\mathbf{l}}, \Tilde{\mathbf{p}}) = \alpha_1\hat{S}_{\mathcal{W}_1}(\hat{\mathbf{f}}) + \alpha_2\hat{S}_{\mathcal{W}_1}(\hat{\mathbf{l}}) + \alpha_3\hat{S}_{\mathcal{W}_1}\left(\Tilde{\mathbf{p}}\right),
\end{equation}
where $\alpha_1, \alpha_2, \alpha_3$ satisfy equation (\ref{eq-4}), $\Tilde{\mathbf{p}}$ is the normalization of $\text{Softmax}(\nicefrac{\mathbf{l}}{T})$, $T$ is the temperature.



\begin{table*}[t]

    \centering
    \caption{Evaluation on the CIFAR-10~\cite{b3} benchmark using ResNet-18 \cite{b5} and WideResNet-28 \cite{b16} as the backbone. 
    }
    \label{tab-1}
    \resizebox{\textwidth}{!}
{
    \begin{tabular}{ccccccccccccccccc}
    \toprule[1.5pt]
    \multirow{3}{*}{\textbf{Model}}&\multirow{3}{*}{\textbf{Methods}} & \multicolumn{14}{c}{\textbf{OOD Datasets}} & \multirow{4}{*}{\textbf{ID ACC}} \\ 
        ~&~& \multicolumn{2}{c}{\textbf{CIFAR-100}} & \multicolumn{2}{c}{\textbf{TIN}} & \multicolumn{2}{c}{\textbf{MNIST}} & \multicolumn{2}{c}{\textbf{SVHN}} & \multicolumn{2}{c}{\textbf{Texture}} & \multicolumn{2}{c}{\textbf{Place365}} & \multicolumn{2}{c}{\textbf{Average}} & ~ \\
         ~&~ & \textbf{FPR@95 $\downarrow$} & \textbf{AUROC $\uparrow$} & \textbf{FPR@95 $\downarrow$} & \textbf{AUROC $\uparrow$} & \textbf{FPR@95 $\downarrow$} & \textbf{AUROC $\uparrow$} & \textbf{FPR@95 $\downarrow$} & \textbf{AUROC $\uparrow$} & \textbf{FPR@95 $\downarrow$} & \textbf{AUROC $\uparrow$} & \textbf{FPR@95 $\downarrow$} & \textbf{AUROC $\uparrow$} & \textbf{FPR@95 $\downarrow$} & \textbf{AUROC $\uparrow$} & ~\\
        \midrule
        \multirow{8}{*}{\rotatebox[origin=c]{90}{ResNet-18 \cite{b5}}}&MSP \cite{b2}& 59.82 & 86.73 & 47.33 & 88.64 & 
        19.22 & 93.95 & \textbf{23.82} & 91.68 & 40.20 & 89.13 & 41.67 & 89.35 & 38.68 & 89.91 & 95.22 \\
        ~&ODIN \cite{b10} & 84.87 & 79.55 & 84.59 & 81.18 & 
        \textbf{15.39} & 
        \textbf{96.60} & 69.04 & 84.62 & 83.19 & 
        83.86 & 76.62 & 83.87 & 68.95 & 84.95 & 95.22 \\
        ~&MDS \cite{b8} & 95.83 & 49.85 & 94.68 & 49.60 & 
        65.67 & 
        66.86 & 44.56 & 84.38 & 98.37 & 
        47.82 & 91.02 & 62.76 & 81.69 & 60.21 & 95.22 \\
        ~&EBO \cite{b3} & 72.69 & 85.55 & 63.68 & 88.35 & 15.49 & 96.32 & 29.34 & \textbf{92.60} & 60.43 & 88.63 & 56.38 & 89.63 & 49.67 & 90.18 & 95.22 \\  
        ~&Gram \cite{b19} & 95.83 & 49.85 & 94.68 & 49.60 & 65.67 & 66.86 & 44.56 & 84.38 & 98.37 & 47.82 & 91.02 & 62.76 & 81.69 & 60.21 & 95.22 \\ 
        ~&KLM \cite{b9} & 92.23 & 77.48 & 81.97 & 80.02 & 68.80 & 87.31 & 69.12 & 83.54 & 70.66 & 82.21 & 96.72 & 78.14 & 79.92 & 81.45 & 95.22 \\  
        ~&GEN \cite{b4} & 63.63 & 86.71 & 52.09 & 88.89 & 18.01 & 95.00 & 
        24.08 & 92.18 & 46.12 & 89.36 & 
        46.12 & 
        \textbf{89.74} & 41.68 & 90.31 &
        95.22 \\ 
        ~& OTOD (Ours) & \textbf{45.19} & \textbf{87.75} & \textbf{40.01} & \textbf{89.32} & 23.64 & 93.97 & 24.34 & 92.35 & \textbf{33.41} & \textbf{90.15} & \textbf{40.36} & 88.83 & \textbf{34.49} & \textbf{90.40} & \textbf{95.22} \\
        \midrule
        \multirow{8}{*}{\rotatebox[origin=c]{90}{WideResNet-28 \cite{b16}}}&MSP \cite{b2}& 51.90 & 88.56 & 41.00 & 90.34 & 
        16.07 & 95.01 & 17.59 & 93.85 & 51.87 & 89.01 & 47.56 & 89.53 & 37.67 & 91.05 & 95.81 \\
        ~&ODIN \cite{b10}& 84.12 & 80.55 & 81.61 & 82.34 & 
        17.77 & 
        95.91 & 79.61 & 80.08 & 90.88 & 
        81.39 & 82.61 & 82.59 & 72.77 & 83.81 & 95.81 \\
        ~&MDS \cite{b8}& 53.99 & 85.93 & 44.63 & 88.09 & 
        37.77 & 
        87.47 & \textbf{9.60} & \textbf{97.77} & \textbf{12.10} & 
        \textbf{97.71} & 49.28 & 86.77 & 34.56 & 90.62 & 95.76 \\
        ~&EBO \cite{b3}& 65.27 & 88.10 & 53.56 & 90.55 & \textbf{11.58} & \textbf{97.38} & 18.61 & 94.98 & 70.03 & 88.28 & 60.78 & 89.85 & 46.64 & 91.52 & 95.81 \\  
        ~&Gram \cite{b19}& 81.97 & 67.56 & 69.63 & 76.80 & 51.00 & 81.07 & 24.22 & 95.26 & 94.64 & 73.89 & 75.09 & 74.31 & 66.09 & 78.15 & 95.81 \\ 
        ~&KLM \cite{b9}& 80.71 & 79.32 & 82.76 & 81.28 & 74.03 & 87.59 & 67.28 & 86.82 & 88.42 & 82.08 & 80.38 & 79.66 & 78.93 & 82.79 & 95.81 \\  
        ~&GEN \cite{b4}& 61.88 & 88.44 & 50.94 & 90.71 & 12.73 & 97.06 & 
        17.88 & 94.95 & 65.86 & 88.76 & 
        57.34 & 
        89.98 & 44.44 & \textbf{91.65} &
        95.81 \\ 
        ~&OTOD (Ours) & \textbf{40.46} & \textbf{89.52} & \textbf{33.56} & \textbf{90.98} & 22.47 & 94.14 & 19.37 & 94.38 & 37.72 & 90.04 & \textbf{38.00} & \textbf{90.14} & \textbf{31.93} & 91.53 & \textbf{95.81} \\
        \bottomrule[1.5pt]
    \end{tabular}
}
    
\end{table*}

\begin{table*}[t]

    \centering
    \caption{Evaluation on the CIFAR-100~\cite{b3} benchmark using ResNet-18 \cite{b5} and WideResNet-28 \cite{b16} as the backbone.     
    }
    \label{tab-2}
    \resizebox{\textwidth}{!}{
    \begin{tabular}{ccccccccccccccccc}
    \toprule[1.5pt]
        \multirow{3}{*}{\textbf{Model}}&
        \multirow{3}{*}{\textbf{Methods}} & \multicolumn{14}{c}{\textbf{OOD Datasets}} & \multirow{4}{*}{\textbf{ID ACC}} \\ 
       ~&~ & \multicolumn{2}{c}{\textbf{CIFAR-10}} & \multicolumn{2}{c}{\textbf{TIN}} & \multicolumn{2}{c}{\textbf{MNIST}} & \multicolumn{2}{c}{\textbf{SVHN}} & \multicolumn{2}{c}{\textbf{Texture}} & \multicolumn{2}{c}{\textbf{Place365}} & \multicolumn{2}{c}{\textbf{Average}} & ~ \\
         ~&~ & \textbf{FPR@95 $\downarrow$} & \textbf{AUROC $\uparrow$} & \textbf{FPR@95 $\downarrow$} & \textbf{AUROC $\uparrow$} & \textbf{FPR@95 $\downarrow$} & \textbf{AUROC $\uparrow$} & \textbf{FPR@95 $\downarrow$} & \textbf{AUROC $\uparrow$} & \textbf{FPR@95 $\downarrow$} & \textbf{AUROC $\uparrow$} & \textbf{FPR@95 $\downarrow$} & \textbf{AUROC $\uparrow$} & \textbf{FPR@95 $\downarrow$} & \textbf{AUROC $\uparrow$} & ~\\
        \midrule
        \multirow{8}{*}{\rotatebox[origin=c]{90}{ResNet-18 \cite{b5}}}&MSP \cite{b2} & 59.12 & 78.54 & 49.98 & 82.22 & 
        63.47 & 73.54 & 55.44 & 79.37 & 61.24 & 78.07 & 55.40 & 79.61 & 57.44 & 78.56 & 77.17 \\
        ~&ODIN \cite{b10} & 60.98 & 78.03 & 55.86 & 81.53 & 50.36 & 82.09 & 62.16 & 75.56 & 59.83 & \textbf{80.46} & 58.51 & 79.80 & 57.95 & 79.58 & 77.17 \\
        ~&MDS \cite{b8} & 89.97 & 52.50 & 81.37 & 58.72 & 
        72.16 & 
        63.88 & 71.10 & 67.55 & 76.00 & 
        73.38 & 83.97 & 58.60 & 79.10 & 62.44 & \textbf{77.33} \\
        ~&EBO \cite{b3} & 58.88 & 79.01 & 52.53 & 82.51 & 57.57 & 77.30 & 50.84 & 82.67 & 60.21 & 79.33 & 56.47 & 79.81 & 56.08 & \textbf{80.11} & 77.17 \\
        ~&Gram \cite{b19} & 91.74 & 50.38 & 91.96 & 51.16 & \textbf{45.17} & \textbf{85.82} & \textbf{22.50} & \textbf{95.22} & 90.30 & 69.27 & 93.69 & 45.01 & 72.56 & 66.14 & 77.17 \\
        ~&KLM \cite{b9} & 88.62 & 74.18 & 70.62 & 79.57 & 63.70 & 71.94 & 58.02 & 80.07 & 75.50 & 76.32 & 83.52 & 76.04 & 73.33 & 76.35 & 77.17 \\  
        ~&GEN \cite{b4} & \textbf{58.56} & \textbf{79.39} & \textbf{49.44} & \textbf{83.28} & 61.42 & 75.66 & 53.68 & 81.53 & 60.78 & 79.53 & \textbf{55.03} & \textbf{80.57} & 56.49 & 79.99 & 77.17 \\ 
        ~&OTOD (Ours) & 59.38 & 78.91 & 50.39 & 83.04 & 56.96 & 76.75 & 50.47 & 81.99 & \textbf{58.38} & 79.36 & 55.24 & 80.08 & \textbf{55.14} & 80.02 & 77.17 \\
        \midrule
        \multirow{8}{*}{\rotatebox[origin=c]{90}{WideResNet-28 \cite{b16}}}&MSP \cite{b2} & \textbf{55.70} & 80.45 & 49.58 & 83.01 & 
        50.50 & 79.61 & 54.58 & 81.61 & 66.18 & 75.86 & 57.49 & 79.69 & 55.67 & 80.04 & 80.07 \\  
        ~&ODIN \cite{b10} & 59.11 & 79.62 & 57.56 & 81.31 & 47.28 & \textbf{83.15} & 72.03 & 72.57 & 66.23 & 77.88 & 64.07 & 79.24 & 61.05 & 78.96 & 80.07 \\
        ~&MDS \cite{b8} & 74.32 & 75.04 & 55.47 & 81.83 & 
        71.04 & 
        70.24 & 34.13 & 90.85 & \textbf{38.97} & 
        91.07 & 65.11 & 78.46 & 56.51 & 81.25 & \textbf{80.33} \\
        ~&EBO \cite{b3} & 57.16 & 81.33 & 51.16 & 83.80 & 
        \textbf{44.73} & 84.08 & 53.44 & 83.39 & 66.97 & 76.46 & 60.92 & 79.70 & 55.73 & \textbf{81.46} & 80.07  \\
        ~&Gram \cite{b19} & 87.34 & 64.90 & 66.87 & 75.43 & 56.29 & 80.55 & \textbf{21.70} & \textbf{93.95} & 76.40 & 80.25 & 83.60 & 64.05 & 65.37 & 76.52 & 80.07 \\
        ~&KLM \cite{b9} & 74.16 & 75.76 & 73.19 & 79.61 & 66.40 & 74.41 & 59.34 & 79.82 & 83.12 & \textbf{72.78} & 81.53 & 75.51 & 72.96 & 76.32 & 80.07 \\  
        ~&GEN \cite{b4} & 55.82 & 81.33 & 48.73 & 84.29 & 48.78 & 80.45 & 53.16 & 83.40 & 65.98 & 76.97 & \textbf{57.32} & \textbf{80.78} & \textbf{54.97} & 81.20 & 80.07 \\ 
        ~&OTOD (Ours) & 56.67 & \textbf{81.52} & \textbf{48.88} & \textbf{84.49} & 47.21 & 81.43 & 52.99 & 83.06 & 69.52 & 75.68 & 57.70 & 80.47 & 55.50 & 81.11 & 80.07 \\
        \bottomrule[1.5pt]
    \end{tabular}
    }
\end{table*}

\subsection{Theoretical Guarantee}\label{theory}
In this section, we provide the theoretical guarantee for the feature part of OTOD (the Equation (\ref{eq-1})) approximately. 

\begin{figure}[t]
\centerline{\includegraphics[width=0.87\linewidth]{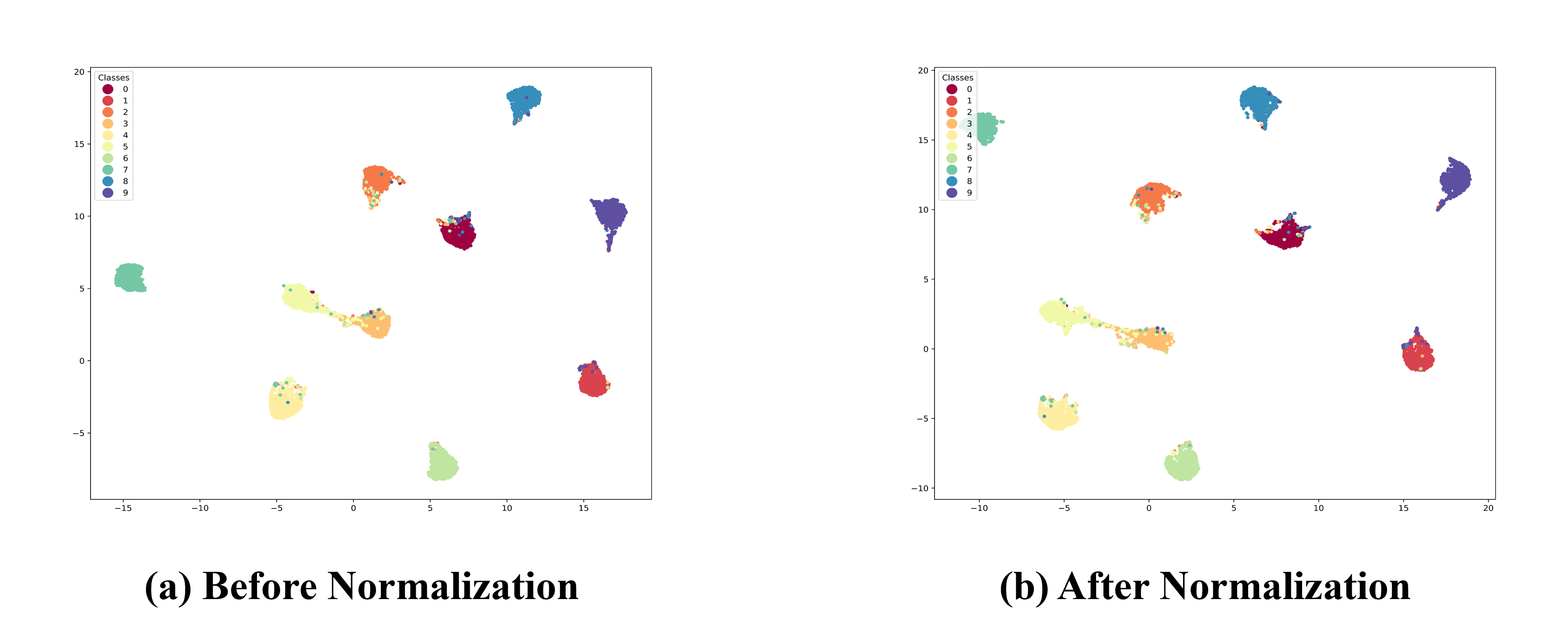}}
\caption{UMAP \cite{b22} visualization of unnormalized (a) and normalized (b) feature distribution using WideResNet-28 \cite{b16} on the CIFAR-10 \cite{b3} benchmark.}
\label{fig-2}
\end{figure}

To this end, we assume that a class conditional distribution in the model's penultimate layer's feature space follows the multivariate Gaussian distribution \cite{b20}, which is empirically demonstrated in \cite{b8}. Herein, we also verify this fact in Fig. \ref{fig-2} (a). In addition, after the normalization operation, the feature distribution does not change markedly (see Fig. \ref{fig-2}(b)). Therefore, we can still use the multivariate Gaussian assumption \cite{b20}, i.e., $\hat{\mathbf{f}}\vert y_i\sim \mathcal{N}(\boldsymbol{\mu}_i, \Sigma),$ where $\boldsymbol{\mu}_i$ is the mean of class $y_i, i=1,\cdots, K$, $\Sigma$ is the covariance matrix. 

Then, we define the Mean Discrepancy (MD) metric of a given score function, which is a concept borrowed from Maximum Mean Discrepancy (MMD) \cite{b21} that quantifies how well OTOD discerns ID samples from the OOD ones. The definition can be stated as follows:

\begin{definition}[Mean Discrepancy]
Denote $S: \Tilde{X}\rightarrow \mathbb{R}$ as a score function, where $\Tilde{X}$ is the model's penultimate layer's feature space. Then, let $P^{in}_{\Tilde{X}}$ be the in-distribution marginal probability distribution on $\Tilde{X}$, $P^{ood}_{\Tilde{X}}$ be the OOD distribution. Therefore, the Mean Discrepancy (MD) for score function $S$ can be written as 
\begin{equation*}
    \mathbb{E}_{\mathbf{x}\sim P^{in}_{\Tilde{X}}} \left[S(\mathbf{x})\right] - \mathbb{E}_{\mathbf{x}\sim P^{ood}_{\Tilde{X}}} [S(\mathbf{x})].
\end{equation*}
\end{definition}

Assume that the out-of-distribution data distribution $P_{\Tilde{X}}^{ood}=\mathcal{N}(\boldsymbol{\mu}^{ood}, \Sigma)$ \cite{b20}, herein $\boldsymbol{\mu}^{ood}$ is the mean of OOD data. Then, we can show the following Theorem, which gives the upper bound of Equation (\ref{eq-1})'s MD.

\begin{theorem}[Mean Discrepancy Upper Bound]
    Suppose that the diameter of $\Tilde{X}$ is bounded by $D$, $V(\Tilde{X})$ is the volume of $\Tilde{X}$. Consider the case where $\Tilde{X}$ is the normalized feature space. Then, $\forall 1\leq i\leq K,$ given the setups above, we have the following inequality:
    \begin{align}
        \mathbb{E}_{\hat{\mathbf{f}}\sim P^{in}_{\Tilde{X}}} \left[\hat{S}_{\mathcal{W}_1}(\hat{\mathbf{f}})\right] &- \mathbb{E}_{\hat{\mathbf{f}}\sim P^{ood}_{\Tilde{X}}} \left[\hat{S}_{\mathcal{W}_1}(\hat{\mathbf{f}})\right] \notag\\
        &\leq  2\Tilde{C} D V(\Tilde{X})  \vert\vert \boldsymbol{\mu}_i - \boldsymbol{\mu}^{ood} \vert\vert_{TV}, \label{ineq-1}
    \end{align}
\end{theorem}
where $\vert\vert\cdot\vert\vert_{TV}$ represents the  total variation norm, $\Tilde{C}$ is a constant. 
\begin{proof}
    Denote $\text{MD} = \mathbb{E}_{\hat{\mathbf{f}}\sim P^{in}_{\Tilde{X}}}[\hat{S}_{\mathcal{W}_1}(\hat{\mathbf{f}})] - \mathbb{E}_{\hat{\mathbf{f}}\sim P^{ood}_{\Tilde{X}}}[\hat{S}_{\mathcal{W}_1}(\hat{\mathbf{f}})]$, and the probability density function of $P_{\Tilde{X}}^{in}(\hat{\mathbf{f}})$ and $P_{\Tilde{X}}^{ood}(\hat{\mathbf{f}})$ as $p_{\Tilde{X}}^{in}(\hat{\mathbf{f}}), p_{\Tilde{X}}^{ood}(\hat{\mathbf{f}})$, respectively. Then we have
\begin{equation*}
    \text{MD} \leq \vert \mathbb{E}_{\hat{\mathbf{f}}\sim P^{in}_{\Tilde{X}}}[\hat{S}_{\mathcal{W}_1}(\hat{\mathbf{f}})] - \mathbb{E}_{\hat{\mathbf{f}}\sim P^{ood}_{\Tilde{X}}}[\hat{S}_{\mathcal{W}_1}(\hat{\mathbf{f}})] \vert
\end{equation*}
\begin{align*}
    ~&\leq \int_{\Tilde{X}} \mathcal{W}_{1} (\hat{\mathbf{f}}, \mathbf{u})p^{in}_{\Tilde{X}} (\hat{\mathbf{f}}) d\hat{\mathbf{f}} + \int_{\Tilde{X}} \mathcal{W}_{1} (\hat{\mathbf{f}}, \mathbf{u})p^{ood}_{\Tilde{X}} (\hat{\mathbf{f}}) d\hat{\mathbf{f}} \\
    &\leq D \left[\int_{\Tilde{X}} \vert\vert \hat{\mathbf{f}} - \mathbf{u} \vert\vert_{TV}\cdot p^{in}_{\Tilde{X}} (\hat{\mathbf{f}}) d\hat{\mathbf{f}}\right. \\
    &+ \left.\int_{\Tilde{X}} \vert\vert \hat{\mathbf{f}} - \mathbf{u} \vert\vert_{TV}\cdot p^{ood}_{\Tilde{X}} (\hat{\mathbf{f}}) d\hat{\mathbf{f}} \right] \\
    &\leq 2D \int_{\Tilde{X}} \vert\vert \hat{\mathbf{f}} - \mathbf{u} \vert\vert_{TV} d\hat{\mathbf{f}} \\
    &= 2D V(\Tilde{X}) \vert\vert  \hat{\mathbf{f}}^* - \mathbf{u}\vert\vert_{TV} \\
    &\leq 2\Tilde{C} D V(\Tilde{X})\vert\vert \boldsymbol{\mu}_i - \boldsymbol{\mu}^{ood} \vert\vert_{TV},
\end{align*}  
where $\hat{\mathbf{f}}^*\in \Tilde{X}$, $\Tilde{C}$ is a constant.
\end{proof}

Notice that, as $\vert\vert \boldsymbol{\mu}_i-\boldsymbol{\mu}^{ood}\vert\vert_{TV}\rightarrow 0$, the right-hand side of the inequality (\ref{ineq-1}) approaches to 0, which implies that the performance of the feature part of OTOD (the Equation (\ref{eq-1})) decreases as $\vert\vert \boldsymbol{\mu}_i-\boldsymbol{\mu}^{ood}\vert\vert_{TV}$ approaches to 0. In other words, Equation (\ref{eq-1}) can detect pronounced distribution shifts.

\section{Experimental Results}
\subsection{Experimental Setup}
\noindent\textbf{Datasets.}\quad In our research, we utilize two common benchmarks to evaluate our methods: the CIFAR-10 \cite{b3} and CIFAR-100 \cite{b3} benchmarks. In the CIFAR-10 benchmark, the OOD datasets used for evaluation are CIFAR-100 \cite{b3}, TIN \cite{b11}, MNIST \cite{b12}, SVHN \cite{b13}, Texture \cite{b14}, Places365 \cite{b15}. In the CIFAR-100 \cite{b3} benchmark, the OOD datasets used for evaluation are CIFAR-100 \cite{b3}, TIN \cite{b11}, MNIST \cite{b12}, SVHN \cite{b13}, Texture \cite{b14}, Places365 \cite{b15}, respectively.

\noindent\textbf{Evaluation metrics.}\quad We use the following metrics to evaluate the OOD detection performance: (\romannumeral 1) the False Positive Rate (FPR) at $95\%$ True Positive Rate; (\romannumeral 2) the Area Under the Receiver Operating Characteristic curve (AUROC); (\romannumeral 3) the in-distribution classification accuracy (ID ACC).

\noindent\textbf{Implementation details.}\quad  
In the experiments, we use ResNet-18 \cite{b5} and WideResNet-28 \cite{b16} as backbones to perform OOD detection on the aforementioned datasets. To ensure fairness, all post-hoc methods use the same pre-trained model and are evaluated on one NVIDIA V100 GPU with Python 3.8.19, CUDA 11.3 + Pytorch 1.13.1. On the CIFAR-100 benchmark \cite{b3}, the temperature of our method is set to $T=3$ for both ResNet-18 \cite{b5} and WideResNet-28 \cite{b16} architectures. On the CIFAR-10 benchmark \cite{b3}, the temperature of our method is set to $T=10$ for both ResNet-18 \cite{b5} and WideResNet-28 \cite{b16} architectures. For both ResNet-18 \cite{b5} and WideResNet-28 \cite{b16} architectures, $\alpha_1, \alpha_2, \alpha_3$ in equation (\ref{eq-3}) are all fixed to $\frac{1}{3}$ on the CIFAR-10/100 \cite{b3} benchmarks. The resolution of the pre-trained model is $32\times 32$ and the training epoch is set to 100. The learning rate of the pre-trained model is taken as $10^{-1}$ using SGD \cite{b17} as the optimizer. Our code is developed based on OpenOOD \cite{b18,b24}.

\noindent\textbf{Baselines.}\quad We compare OTOD with seven post-hoc OOD detection baselines, including MSP \cite{b2}, ODIN \cite{b10}, MDS \cite{b8}, EBO \cite{b3}, Gram \cite{b19}, KLM \cite{b9}, GEN \cite{b4}. We use OpenOOD \cite{b18,b24} to implement all the aforementioned methods.  For ODIN \cite{b10}, the temperature is set to $T=1000$ following the original work. For EBO \cite{b3}, the temperature $T$ is set to $1$.

\begin{table}[t] 
\begin{center}
\caption{Ablation of the input of OTOD on the CIFAR-100 benchmark using ResNet-18 as the backbone and the temperature $T$ is taken as 3. Here we report the mean value of FPR and AUROC over 6 OOD datasets.} 
\label{tab-ab}
    \resizebox{0.45\textwidth}{!}{
\begin{tabular}{c|ccc|c|c}
  \toprule[1.5pt]
    & \makecell[c]{Features} & \makecell[c]{Logits} & \makecell[c]{Probs} & \textbf{FPR@95}$\downarrow$ & \textbf{AUROC} $\uparrow$
  \\
  \midrule
  (\romannumeral 1)&  \Checkmark & \XSolidBrush & \XSolidBrush & 62.98 & 74.54 \\
  (\romannumeral 2)&  \Checkmark  & \Checkmark & \XSolidBrush  &  58.68  & 77.98 \\
  (\romannumeral 3)&  \Checkmark & \Checkmark & \Checkmark & \textbf{55.14} & \textbf{80.02} \\

  \bottomrule[1.5pt]
\end{tabular}
}
\end{center}
\end{table}

\subsection{OOD Detection Performance Results}
\noindent\textbf{Evaluation on CIFAR-10.}\quad 
The comparison results on the CIFAR-10 \cite{b6} benchmark are given in Table \ref{tab-1}. As shown in the first half of Table \ref{tab-1}, OTOD outperforms all the other scoring functions on various OOD datasets by a considerable margin using ResNet-18 as the backbone. Furthermore, it is noteworthy that OTOD surpasses GEN by $12.71\%$ FPR@95 on Texture. From the second half of Table \ref{tab-1}, we can find that OTOD surpasses GEN by $12.51\%$ on FPR@95 averaged on six OOD datasets taking WideResNet-28 as the backbone. These facts indicate that OTOD works for both ResNet-18 and WideResNet-28 architectures on the CIFAR-10 benchmark.

\noindent\textbf{Evaluation on CIFAR-100.}\quad The results on CIFAR-100 benchmark is given in Table \ref{tab-2}. From the first half of Table \ref{tab-2}, we can find that OTOD achieves the best mean FPR@95 compared to other methods. Moreover, OTOD outperforms KLM by $18.19\%$ FPR@95 and $3.67\%$ AUROC on average. As shown in Table \ref{tab-2}, OTOD also achieves competitive results compared with GEN. In the second half of Table \ref{tab-2}, we can see that OTOD also works well using WideResNet-28 as the backbone. 

\subsection{Ablation Analysis of OTOD Input}
In this section, our purpose is to investigate the effects of each input in OTOD on the CIFAR-100 benchmark taking ResNet-18 as the backbone. By comparing setting (\romannumeral 1), (\romannumeral 2), and (\romannumeral 3) in Table \ref{tab-ab}, we can observe that the combination of features, logits and softmax probability inputs enhances the performance of OTOD scoring function. Note that, the AUROC of setting (\romannumeral 1) is $5.48\%$ lower than the AUROC of setting (\romannumeral 3).

\subsection{The Effect of Hyperparameters}
\begin{figure}[t]
\centerline{\includegraphics[width=0.9\linewidth]{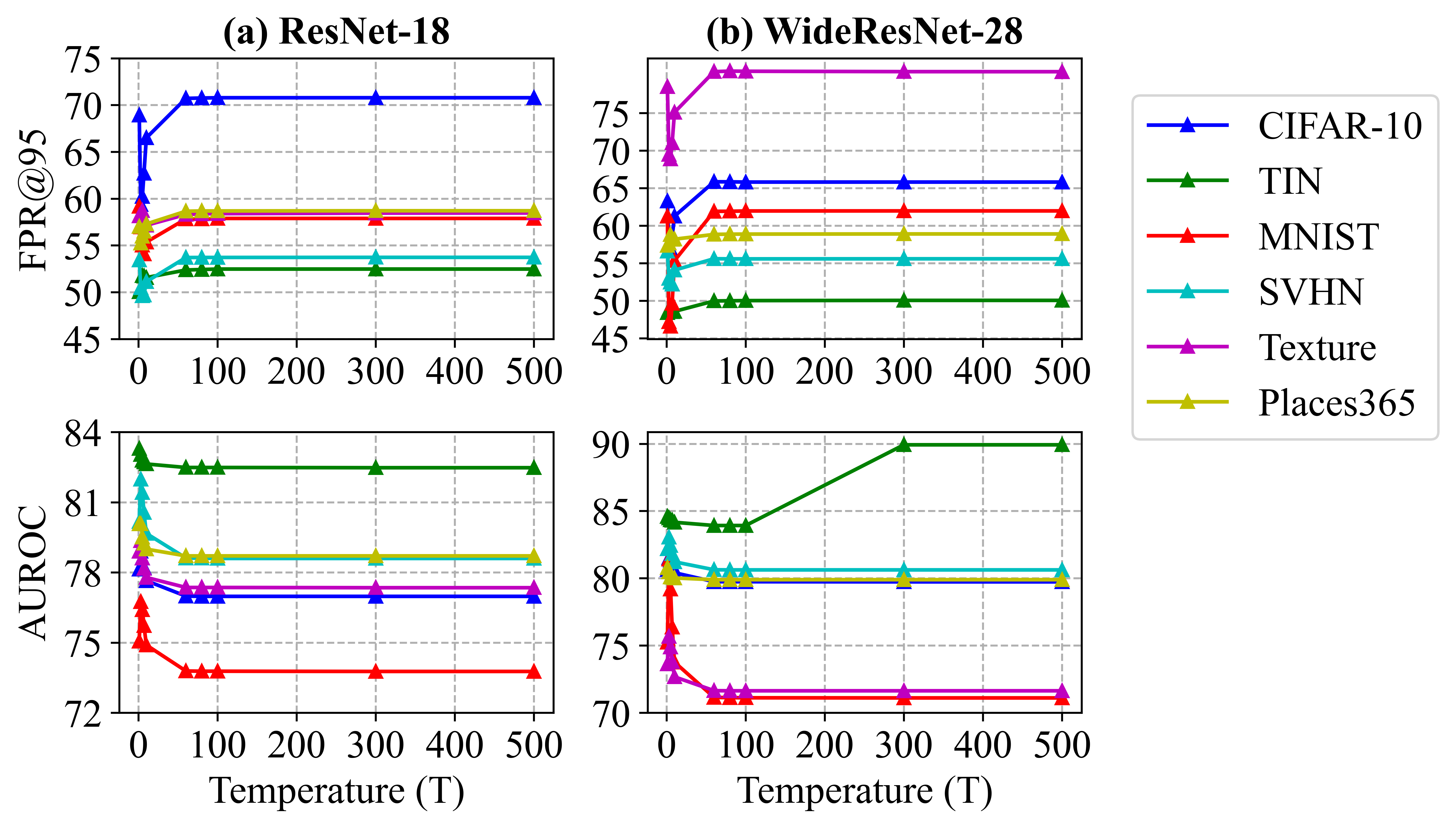}}
\caption{(a)(b) Hyperparameter analysis on temperature $T$ using ResNet-18 and WideResNet-28 as backbones on the CIFAR-100 benchmark.}
\label{fig-3}
\end{figure}

 In this section, we give a detailed analysis of the temperature $T$ in OTOD on the CIFAR-100 benchmark taking ResNet-18 and WideResNet-28 as backbones. The results are shown in Fig. \ref{fig-3}. From Fig. \ref{fig-3}, we observe that when temperature $T$ is small, increasing $T$ can enhance the OOD detection performance on both ResNet-18 and WideResNet-28 architectures. While $T$ is large, increasing $T$ will harm the OOD detection results on both ResNet-18 and WideResNet-28 architectures.

\section{Conclusion}
In this study, we introduce a novel scoring function, named OTOD, for detecting OOD samples. We calculate OOD scores for test samples using Wasserstein-1 distances, which do not require the use of any in-distribution data during the OOD testing phase. Afterward, We leverage information from feature, logits, and softmax probability spaces to achieve better OOD detection performance using our optimal transport-based score function. Moreover, we apply the temperature scaling technique to further improve the detection result. Both theoretical and experimental results demonstrate the superior performance of our score function.

\end{document}